# Miti360: A Comprehensive Dataset for Improved Reforestation Monitoring

Cedric Kiplimo[1*], Samuel Mbatia[1], Ciira wa Maina[1], Arthur Sichangi[2], Dennis Gitundu[2]

[1]Centre for Data Science and Artificial Intelligence (DSAIL), Dedan Kimathi University of Technology, Nyeri, Kenya
[2]Institute of Geomatics, GIS and Remote Sensing (IGGReS), Dedan Kimathi University of Technology, Nyeri, Kenya
[*]Correspondence: cedric.kiplimo@dkut.ac.ke

**Abstract**

Over the past decade, interest in applying machine learning (ML) to automate forest monitoring has grown significantly. However, existing training datasets are predominantly drawn from North America, Europe, Asia, and Australia, leaving a critical gap in African forestry data. To address this limited geographic diversity, we present Miti360, a comprehensive dataset for reforestation monitoring that comprises high-resolution imagery, ground truth data, and longitudinal weather data. Data collection occurred within a 770-ha reforested section of the Kieni Forest in Kenya between March 2023 and February 2025. Miti360 comprises aerial photos (orthophotos and tiles) with tree bounding box annotations, terrestrial images (single and stereo), and detailed data records including tree biophysical parameters, species, and GPS coordinates, alongside historical weather data. Aerial surveys utilized a DJI Mavic 2 Pro, with imagery stitched via Agisoft Metashape and tiled using ArcGIS Pro, while terrestrial captures used smartphones and custom stereo cameras. Miti360 enables the training of ML systems for tasks such as accelerating tree censuses, matching species to geographical areas, modelling growth based on weather conditions, and developing digital twin frameworks. Models can be trained on Miti360 to address challenges specific to Sub-Saharan Africa, ultimately advancing reforestation monitoring and fostering sustainable forestry practices in underrepresented regions. We demonstrate the utility of this dataset by successfully tracking tree crowns across three years and improving the DeepForest model's box precision and box recall by 12% and 69% respectively through fine-tuning on Miti360.

## 1 Introduction

Forest monitoring is essential in protecting ecosystems, mitigating climate change, and securing local livelihoods [1], [2]. In today's age of Artificial Intelligence (AI), Machine Learning (ML) has the potential to improve forest monitoring efforts when coupled with appropriate data such as satellite images, aerial photos, and ground measurements [3], [4], [5]. The application of these tools to analyse satellite and drone imagery can speed up the accurate estimation of biomass and individual tree monitoring [6], [7], [8], [9], [10], [11], [12].. Major forestry datasets used for ML training include NEON Crowns [3] and the Auto Arborist dataset [4] with data from North America, ReforesTree [13] with data from Ecuador, and a dataset from Northern Australia [14]. Despite the huge interest in applying ML to forestry, the scarcity of machine learning ready datasets related to forestry in Africa remains a major hindrance [5], [15]. The number of African datasets on forestry is growing but remains far behind those from developed nations [16], [17]. We undertook the data collection and curation to address the existing data gap by collecting, annotating, and providing access to information that supports data-driven decision-making in establishing reforested stands and monitoring the success of reforestation efforts.

In practice, data on forest resources are collected systematically through a process called forest inventory [1], [18], [19], [20], [21]. This process is crucial for understanding forest composition, structure, and health, thereby supporting effective management and conservation efforts. Accurate forest monitoring, especially at the individual tree level, will be immensely beneficial to African countries where forest inventories still lag those of developed nations in terms of scale, accuracy and reliability [1], [22], [23], [24]. As an example, Kenya faces serious ongoing deforestation challenges, with studies of key montane ecosystems revealing that 21.9% of the Mau Forest Complex — some 88,493 ha — was lost at an annual rate of about 0.82% per year between 1986 and 2017, with agriculture as the dominant driver, while an encouraging 18.6% of the disturbed forest area is now undergoing recovery [25]. Figure 1 shows an example of a recently reforested site in Kieni Forest as part of Kenya's massive tree planting drive.

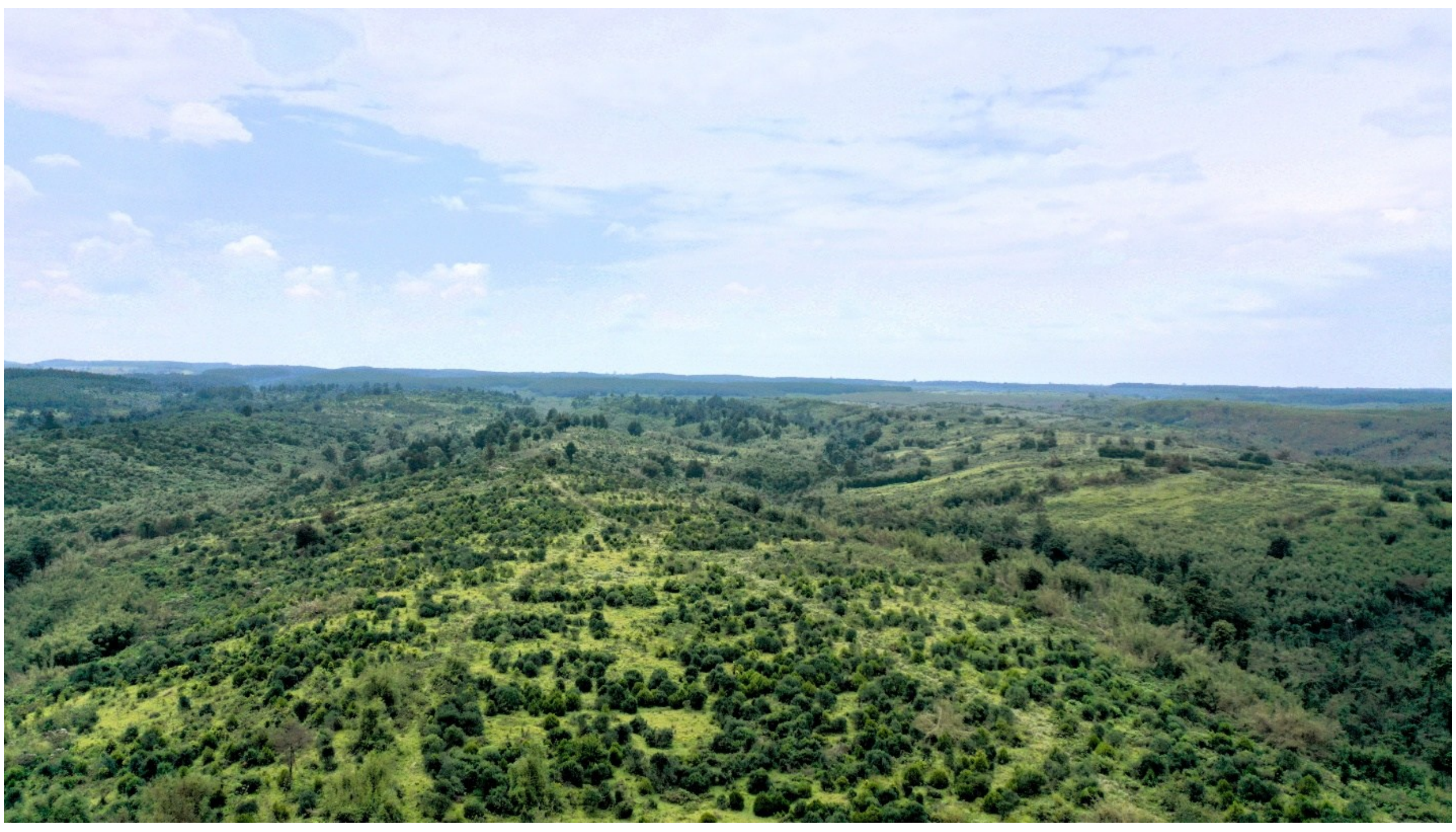

*Figure 1: Oblique aerial view of a section of the study area, displaying a mosaic of early-stage saplings and scattered, mature residual trees. Image acquired on 24 February 2026.*

The current efforts to plant up to 15 billion trees by 2032 [26] present the unique challenge of monitoring the growth of those individual trees to ascertain their survival and assess their growth rates. The Miti360 dataset provides the necessary empirical foundation for developing and benchmarking machine learning models designed to address these challenges. Similar efforts have been undertaken in the Global North [4], [27], [28], [29].

In recent years, the digital twin technology has emerged as a compelling approach for monitoring processes reliably, particularly in manufacturing and industrial processes where digital systems are deployed to automate, optimise and enhance productivity [30], [31], [32]. Forest digital twins are also emerging as a technological advancement in forest monitoring [33], [34]. These digital twins are real-time data-driven replicas of physical forest ecosystems that enable continuous monitoring, simulation and decision support for forest management and conservation support [33], [35], [36]. The development of these digital twins requires high quality data from real world forest ecosystems with data modalities such as satellite imagery, LiDAR point clouds, and IoT sensor data commonly used [37], [38]. As such, comprehensive datasets are needed to develop theses forest digital twins, particularly those related to monitoring reforestation efforts in Sub-Saharan Africa and the Global South [33], [34], [38], [39].

We present Miti360, a comprehensive dataset that brings together high-resolution aerial photos, terrestrial images, ground collected data (biophysical parameters, species information, and GPS locations), and historical weather data. Our objective was to create a dataset which would facilitate an understanding of the effectiveness of reforestation efforts through the analysis of both forest structure and the impact of weather on the growing trees with machine learning. To harness the full potential of data-driven forestry and address pressing challenges in reforestation monitoring, advanced machine learning techniques are being integrated into research and practical field applications. In particular, the machine learning algorithms trained using Miti360 will be able to, among other things:

i. Identify and classify species of individual young trees, saplings and seedlings in heterogeneous stands from drone images,
ii. Estimate biophysical parameters such as crown diameter and tree height and from these derive biomass estimates, and
iii. Predict changes in tree biophysical parameters and stand volume in relation to prevailing weather conditions and tree species.

We believe that this will make accurate quantitative analysis of reforested stands possible and this will in turn promote reforestation efforts.

The remainder of this paper is structured as follows: Section 2 (Methods) details the data acquisition and processing framework. Section 3 (Data Record) outlines the contents of the dataset and how to access them, while section 4 (Dataset Overview) provides a high-level summary of the statistical distribution of the dataset contents. Section 5 (Technical Validation) presents the sample experiments performed with the dataset to demonstrate its use cases. Finally, section 6 (Usage Notes) provides the research community with some usage guidelines. A preliminary description of the metadata schema and initial data collection phase was previously presented as a conference abstract at the American Geophysical Union Fall Meeting 2025 [40].

## 2 Methods

### 2.1 Study Area

The study area is a reforested stand within the Kieni Forest in Kenya (Figure 2).

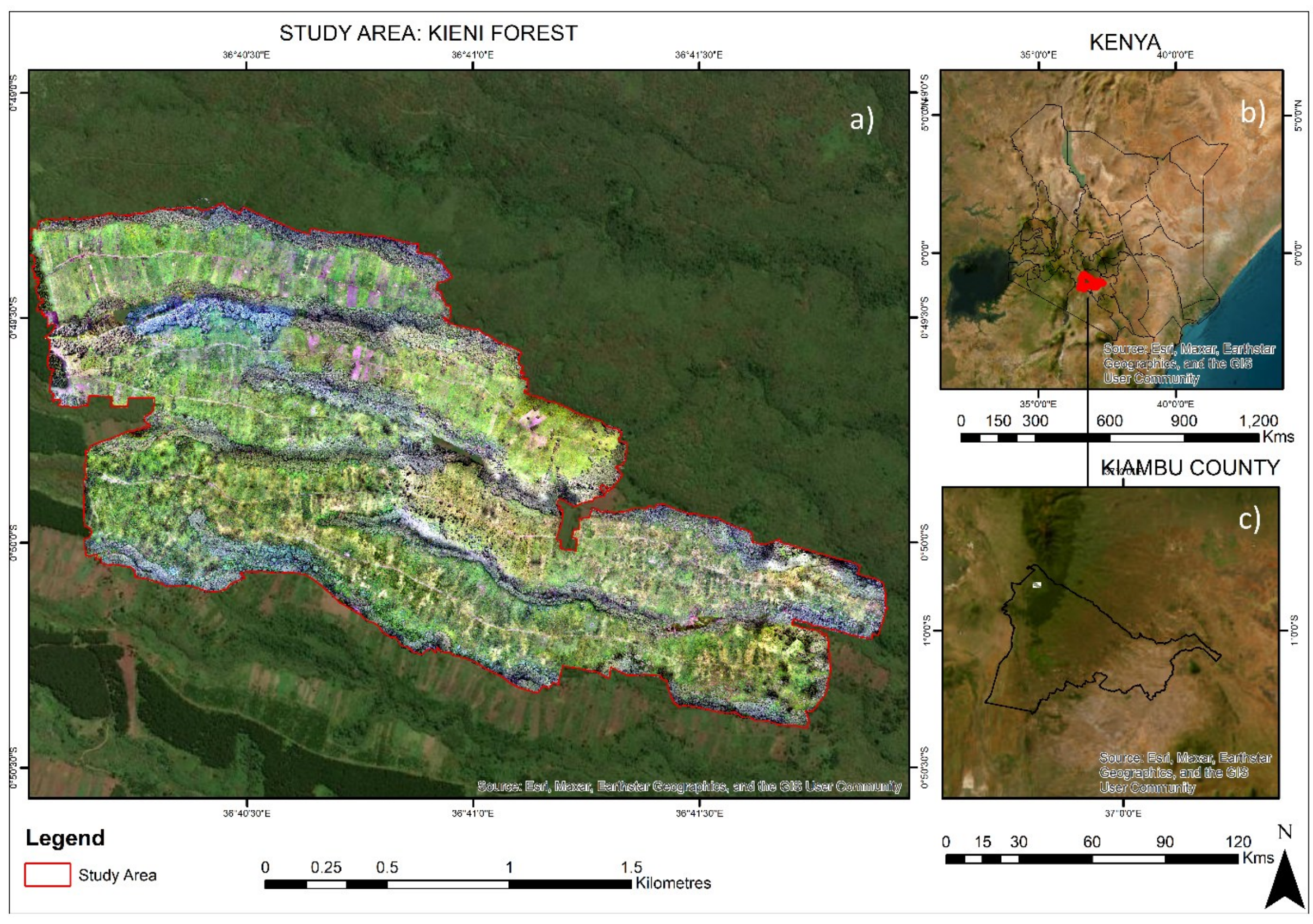

*Figure 2: Location of the study area. a) High-resolution satellite imagery showing the reforested section of Kieni Forest outlined in red. Regional and localized spatial contexts are provided by the inset maps, which pinpoint the site's position within b) Kenya and c) Kiambu County.*

This forest is situated within the Aberdare Ecosystem, one of Kenya's five main water towers which supplies 80% of the water used in Nairobi [41]. The entire ecosystem spreads across four counties—Kiambu, Murang'a, Nyeri, and Nyandarua—with Kieni Forest located within Kiambu. Kieni Forest lies between 2200 m and 2684 m above sea level and receives rainfall of 1150 mm to 2560 mm annually [42]. The long rain season stretches from March to June, while short rains are received between October and December. Its soils are rich in organic matter, making them fertile and favourable to developing thick undergrowth. Its vegetation comprises natural forests, plantations, bamboo, meadows, and tea zones. The replanted section covers an area of 770 ha and contains more than ten indigenous species, mainly *Dombeya torrida*, *Juniperus procera*, *Olea africana*, and *Prunus africana*. At the beginning of this data collection, the crown diameters range from 0.35 m to 6.6 m, the heights from 0.9 m to 7.2 m, and the basal diameters from 2.2 cm and 32.8 cm. Being a reservoir of biodiversity, Kieni Forest also has a wide range of fauna, such as the African elephant (*Lexodonta africana*), duiker (*Neottragus moschatus*), Bush pig (*Patomochoerus porcuso*), and mongoose (*Helogale parvula*), among others [42], [43].

## 2.2 Materials

The aerial images were captured using a DJI Mavic 2 Pro equipped with a specialized Hasselblad L1D-20c gimbal camera. This system utilizes a large 1-inch CMOS sensor capable of capturing 20-megapixel still images at a maximum resolution of 5472 x 3648 pixels. The lens provides a wide-angle perspective with a focal length of 10.26mm (roughly a 28 mm equivalent in 35mm format) and an adjustable aperture range of f/2.8 to f/11. For these specific captures, the camera was set to an ISO of 100 with an exposure time of 1/240 seconds at f/5, utilizing a 24-bit depth for rich sRGB colour representation.

Ground Control Points (GCPs) were collected for georefencing the aerial images. These were collected using both South and Kolida RTK GNSS receivers (during the different phases of data collection), leveraging their high-precision multi-constellation tracking capabilities across GPS, GLONASS, BeiDou, and Galileo satellites. These survey-grade units are engineered with a 1598-channel GNSS engine and advanced Linux-based operating systems, achieving a horizontal positioning precision of 8 mm + 1 ppm and a vertical precision of 15mm + 1 ppm in Real-Time Kinematic (RTK) mode. Designed for rugged fieldwork, the receivers feature an IP67/IP68 waterproof rating and integrated Inertial Measurement Units (IMU) for tilt compensation up to 60°.

The stereo camera used in the data collection is an inhouse camera comprising two Logitech C270 USB web cameras, which have a resolution of 720 × 1280 pixels, in an assembly with a baseline of 12.9 cm. This stereo camera was powered by the TreeVision software [44] running on an NVIDIA Jetson Nano 2 GB Developer Kit and operated via a connected HDMI mini screen. The features of this kit include a quad-core ARM A57 @ 1.43 GHz processor, 4 GB 64-bit LPDDR4 RAM, and a 128-core Maxwell GPU. The single images were captured using an Oppo F11 smartphone with a 3840 × 2160 camera sensor. To capture the GPS coordinates of each tree, the Garmin GPSMAP 64s handheld outdoor GPS was used. Finally, the ground truth biophysical parameters were taken using a tape measure and a graduated height pole.

## 2.3 Field Data Collection

We undertook the data collection and curation to address the existing data gap by collecting, annotating, and providing access to information that supports data-driven decision-making in establishing reforested stands and monitoring the success of reforestation efforts.

### *2.3.1 Sampling Design*

For this project, a systematic sampling approach was used to define the sampling plots and random sampling employed during ground data collection within each plot. This sampling approach was chosen because it provides a good balance between statistical rigour and operational feasibility in the field. Additionally, it is compatible with remote sensing data such as gridded raster data, thus making geostatistical modelling with collected data possible. The study area was subdivided into contiguous 100 m × 100 m units distributed across the entire study area (Figure 3). Sampling plots were then selected systematically from these units such that the distance between consecutive plots was 100 m horizontally and 200 m vertically. In Figure 3, the coloured squares are the sampling plots. The sampling design yielded a total of 56 plots. Of these, trees were sampled from only 41 (yellow squares) while the rest (red squares) were labelled as invalid plots because they were either covered entirely by bamboo bushes or they had no trees. We planned to randomly sample 15 trees per plot whenever a plot contained at least 15 trees, aiming for 615 trees in total, but some plots had fewer than 10 trees.

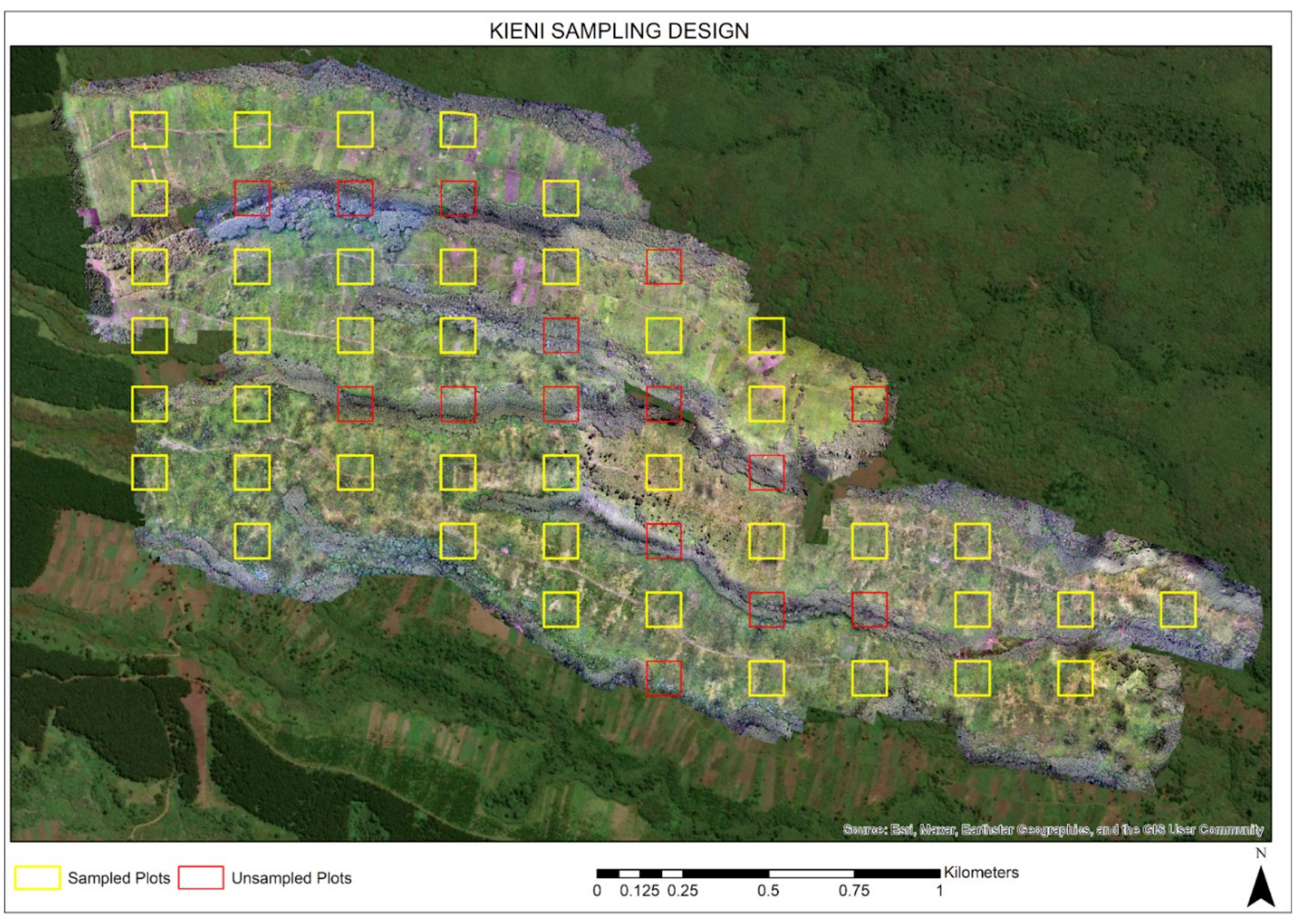

*Figure 3: Sampling design showing the spatial distribution of sampling plots, with sampled plots represented as yellow squares and unsampled plots as red squares*

### *2.3.2 Ground Truth Data Measurement & GPS Data Collection*

The tree biophysical parameters measured during the tree sampling were the tree height (TH), crown diameter (CD), and basal diameter (BD). The TH was measured using a calibrated pole while the other two measurements were obtained using a measuring tape. The location of each sampled tree was recorded using a Garmin GPSMAP 64s handheld GPS receiver with an approximate positional accuracy of 3 m.

### *2.3.3 Terrestrial Image Collection*

For each tree sampled in the study area, two kinds of terrestrial images were captured – stereoscopic (paired left and right) images and single view images. The stereo images were captured using the inhouse stereo camera described in section 2.2 while the single images were taken using the smartphone described in the same section. In both cases, the images were acquired with the target tree centered in the frame and as the primary subject.

## 2.4 Drone Survey

Planning for the drone survey involved two main steps. First, the reforested area was divided into survey blocks that were small enough to be fully covered using a single drone battery. Second, flight plans were prepared for each block using consistent flight parameters.

### *2.4.1 Survey Block Design*

The Area of Interest (AOI) was outlined by digitizing the reforested sections of Kieni Forest using a high-resolution base satellite image in ArcGIS. A grid of square quadrats measuring 150 meters was then overlaid and intersected with the digitized AOI. The quadrats were deliberately kept small to allow for merging rather than splitting during the survey design. Adjacent intersected quadrats were merged to create the final survey blocks. Each block was limited to a maximum area of 0.2 square kilometers, which was suitable for a single drone battery when flying at approximately 90 meters altitude with a flight path spacing of about 30 meters. These parameters were informed by previous flight tests. In total, 21 survey blocks were generated, each measuring between 0.08 and 0.18 square kilometers in size (Figure 4).

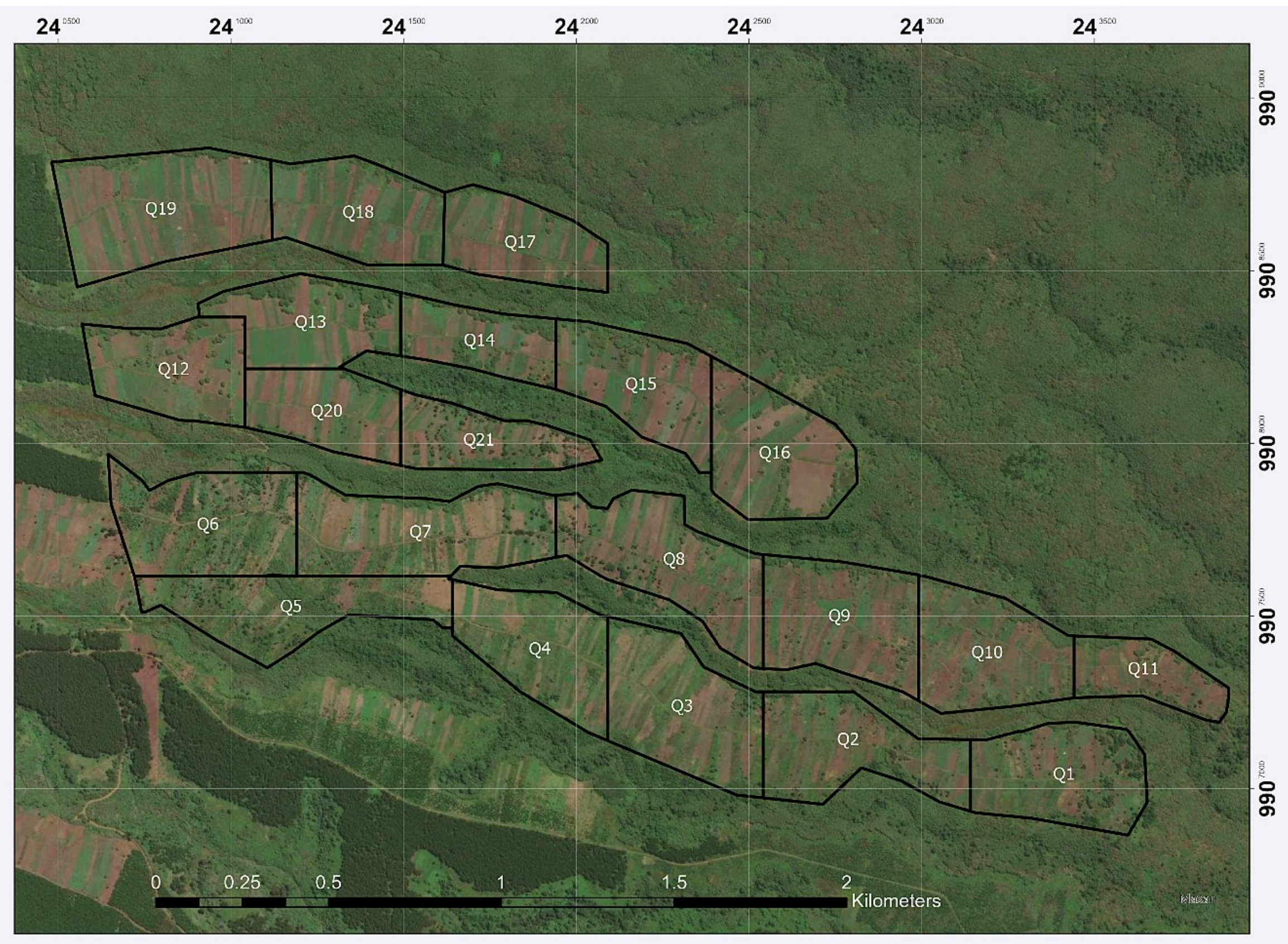

*Figure 4: Final Survey Blocks, numbered according to the order of data collection.*

### 2.4.2 *Ground Control Points*

Ground control points were established by placing X-shaped marks on the ground using chalk dust. These points were distributed in a representative manner across the study area and positioned in locations with unobstructed views of the sky to ensure high visibility for the drone survey.

### 2.4.3 *Flight Planning*

Flight plans were developed for each survey block using DJI Flight Planner software. A ground sampling distance (GSD) of 2 cm was targeted, resulting in a flying altitude of approximately 88 meters (288 feet) above the take-off point. To ensure sufficient image overlap for high-quality orthomosaic generation, a forward overlap of 80% and a side overlap of 70% were applied. These parameters yielded a flight path spacing of approximately 32 meters. The final flight plan (Figure 5) depicts the full mapped area, survey block, flight paths, and the individual camera exposure points.

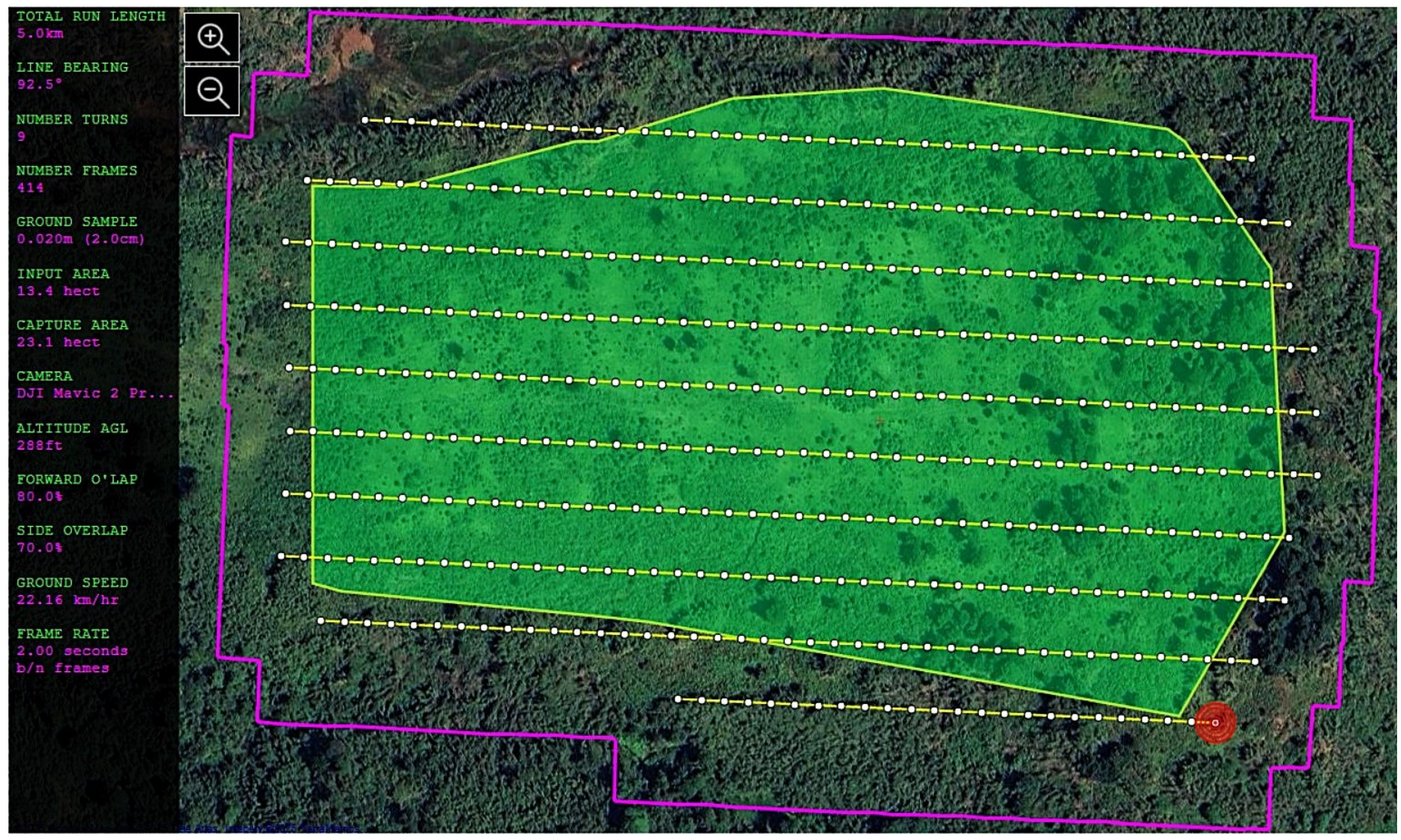

*Figure 5: Sample flight plan showing the full mapped area (purple outline), survey block (green polygon), flight paths (yellow lines), and individual camera exposure points (white dots)*

### *2.4.4 Drone Image Processing*

Drone image processing and georeferencing were executed in Agisoft Metashape using Structure-from-Motion (SfM) techniques on over 9,200 captured images. High forward and side overlaps enabled the software to identify key feature points for initial alignment based on onboard GPS metadata. To ensure geospatial precision, Ground Control Points (GCPs) were incorporated to optimize camera alignment and minimize distortion. A dense point cloud was then generated using multi-view stereo algorithms to create a Digital Elevation Model (DEM). Finally, images were projected onto this DEM to produce a seamless, high-resolution orthophoto, exported in GeoTIFF format for subsequent GIS analysis. More information about the processing can be found in our technical report [45].

## 2.5 Image Annotation

### *2.5.1 Terrestrial Images*

For the terrestrial images (both stereo and single) captured in the study area, two labels were created – a semantic segmentation mask and a species ID. The masks were created using the Labelbox platform, which provides a suite of annotation tools including AI-assisted labelling. Each mask was created through a two-step process. First, AI-assisted labelling powered by the segment anything model (SAM) [46] was used to form a preliminary mask. This was followed, if necessary, by edits to the preliminary mask to make it more accurate. The species of each tree was identified after visual inspection of the image. The annotators could do this with good accuracy after adequate training on species identification had been carried out.

#### *2.5.2 Orthophotos*

As part of the workflow, we conducted detailed data annotation and labelling on orthophoto tiles derived from drone-captured imagery of Kieni Forest. The orthophoto was divided into manageable tiles to facilitate precise annotation of individual tree crowns using Label Studio [47], an open-source data labelling tool. Each tile was carefully examined, and bounding boxes were manually drawn around tree crowns to demarcate their extents. The annotations were saved in JSON format, which stores both the geometric coordinates of the bounding boxes and associated metadata, such as class labels (e.g., tree). The use of Label Studio streamlined the annotation process by providing an intuitive interface for drawing, editing, and exporting labelled data. Quality control measures, including cross-validation between annotators and iterative reviews, were implemented to minimize errors.

### 2.6 Quality Control

Quality control measures were integrated at every stage, starting with daily post-collection verifications to immediately resolve missing attributes or file errors. Prior to labelling, the team underwent comprehensive training on computer vision and specific annotation tools to guarantee high proficiency. During the annotation phase in Labelbox, a strict review workflow was implemented, utilizing consensus scoring where tasks with less than 80% overlap between annotators were rejected for rework. Similar cross-validation protocols were applied to orthophoto bounding boxes to minimize errors. Finally, the completed dataset underwent a rigorous inspection via random sampling to ensure all components were present and met the highest quality standards.

## 3 Data Record

### 3.1 Summary of the Collected and Labelled Data

The data collected in this study consists of aerial and ground-based images, individual tree attributes and aggregated data from TAHMO [48] weather stations. These are captured in Table 1 below.

*Table 1: A tabular summary of the dataset*

| # | Data Category | Data Type | Quantity | Format |
|---|---|---|---|---|
| 1 | Drone Images | Orthophoto | 3 | TIF |
| | | Tiles | 1117 | TIF |

| | | Tree crown annotations | 61866 | JSON |
|---|---|---|---|---|
| | | Tree crown species | 1208 | JSON |
| | | Tree species shapefile | 1208 | SHP |
| 2 | Tree ground measurements | Numeric data | 1208 (604 trees 2 times in 2024 - 2025) | JSON |
| 3 | Ground based single images | Images and tree masks | 1208 (604 trees 2 times in 2024 - 2025) | JPEG |
| 4 | Tree stereo images | Images and tree masks | 2416 (604 trees 2 times in 2024 - 2025) | JPEG |
| 5 | Weather data from 40 stations | Time series data | 8 years daily data | API endpoint |

Figure 6 and Figure 7 show sample image types from the dataset together with their annotations (masks for terrestrial images and tree crown bounding boxes for aerial images). These figures show the image masks for both terrestrial images as well as the bounding boxes around individual trees on the orthophoto tiles.

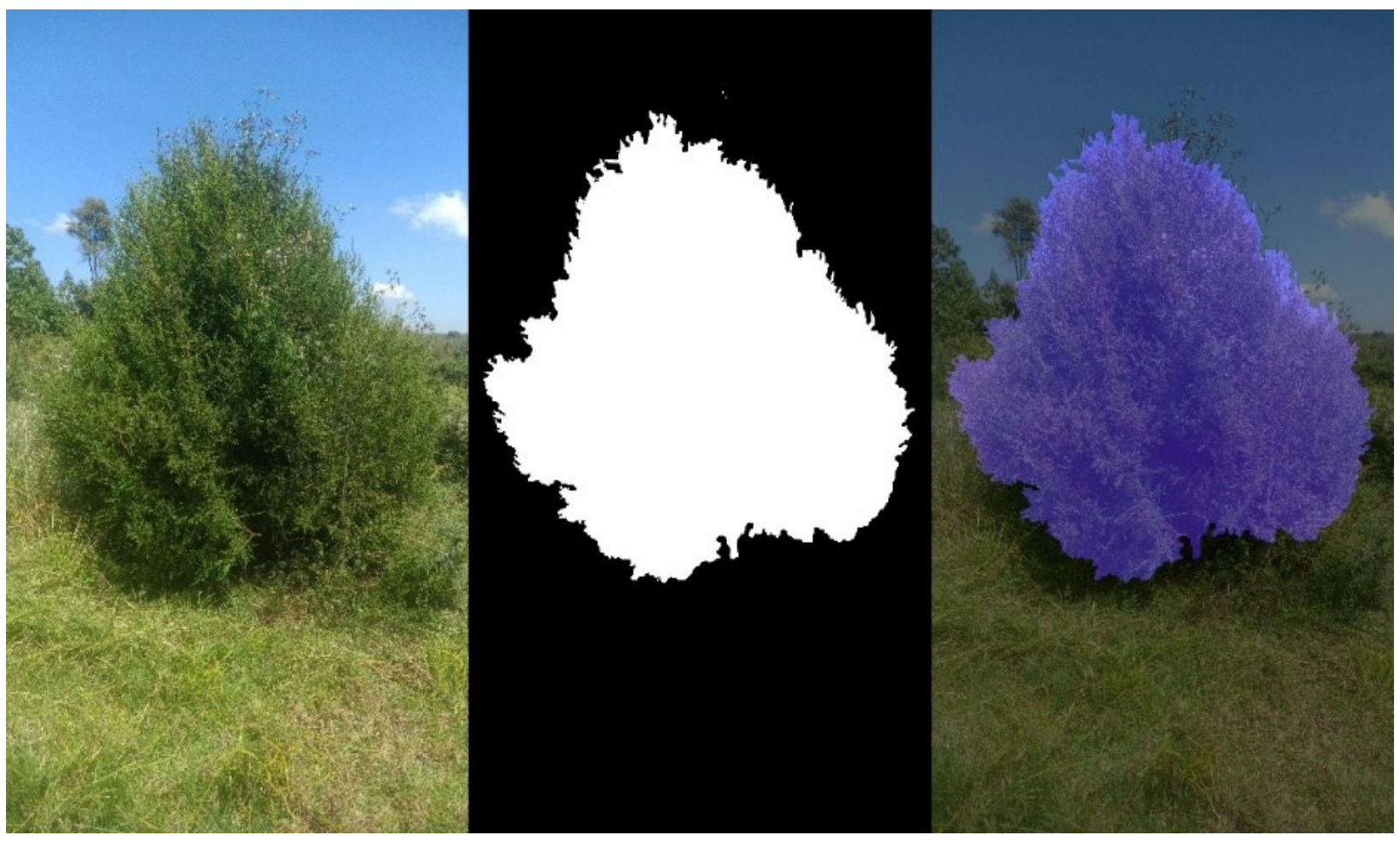

*Figure 6: Sample terrestrial image from the dataset together with its annotation*

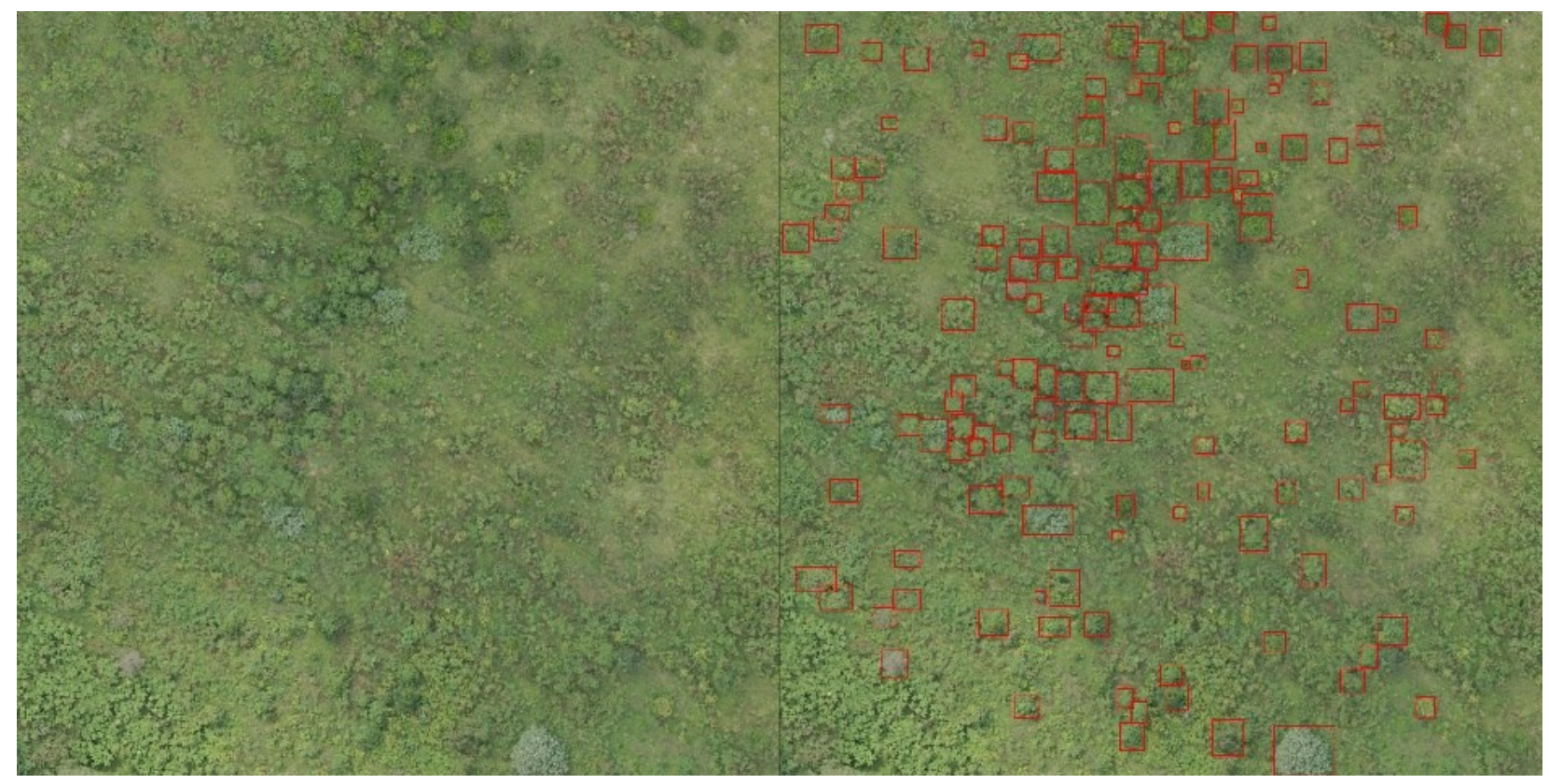

*Figure 7: Sample orthophoto tile from the dataset together with its annotations*

### 3.2 Spatial Distribution of Sampled Trees

The spatial distribution of the sampled trees in the study area is shown in Figure 8.

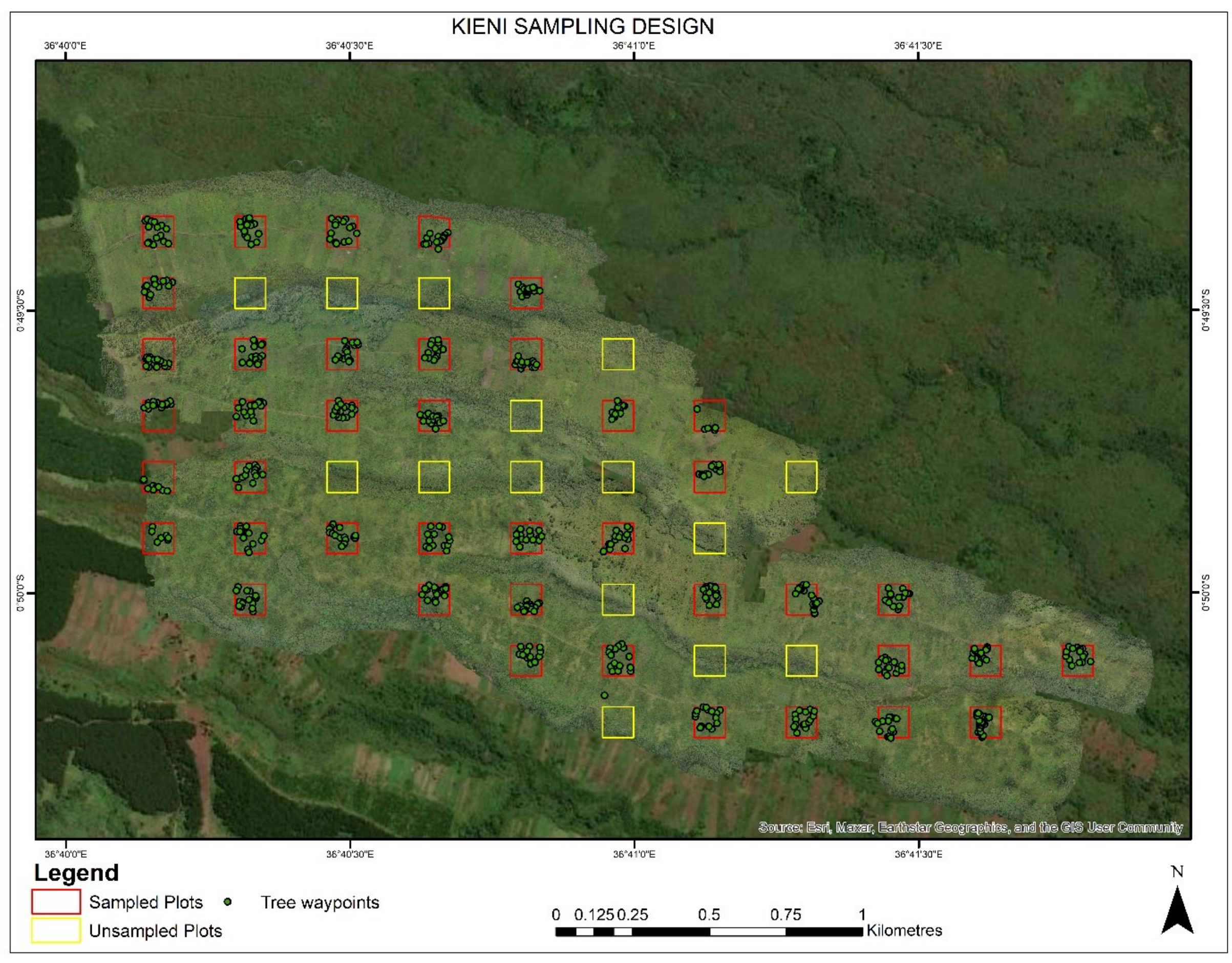

*Figure 8: Map showing the spatial distribution of sampled trees, with tree markers contained in the red squares (sampled plots) and no markers in the yellow squares (unsampled plots)*

### 3.3 Dataset Hosting

All the components of the Miti360 dataset are hosted in publicly accessible Google Cloud Storage buckets. The download links can be found in either the dataset's GitHub repository or the dataset's website.

## 4 Dataset Overview

The distribution of the tree heights, crown diameters, and basal diameters as well as their joint distributions are shown in Figure 9 and Figure 10 respectively.

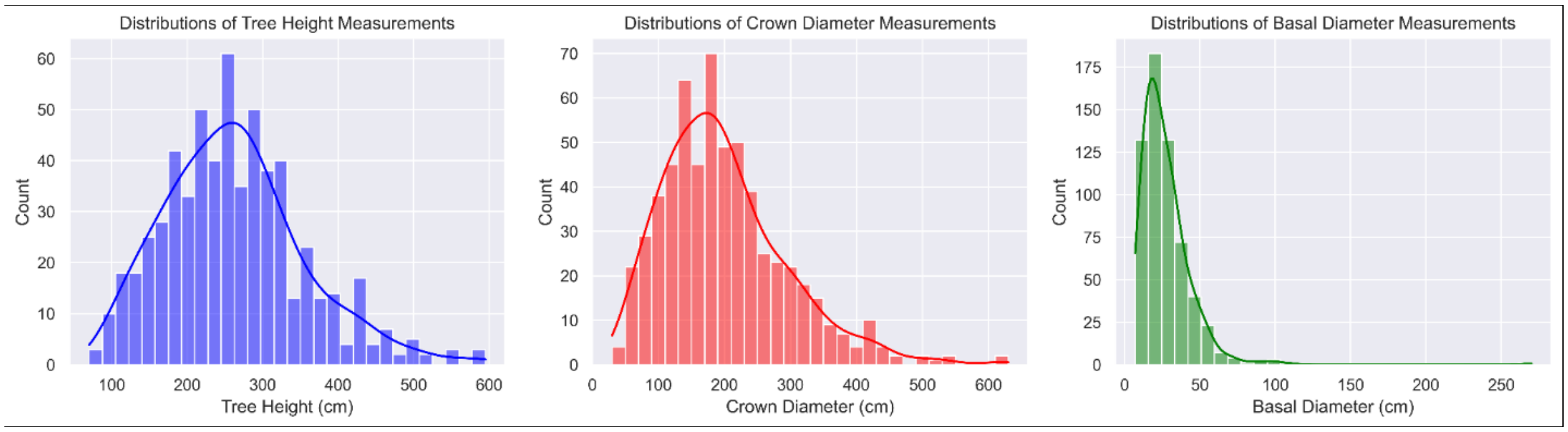

*Figure 9: Distribution of individual tree biophysical parameters in 2024*

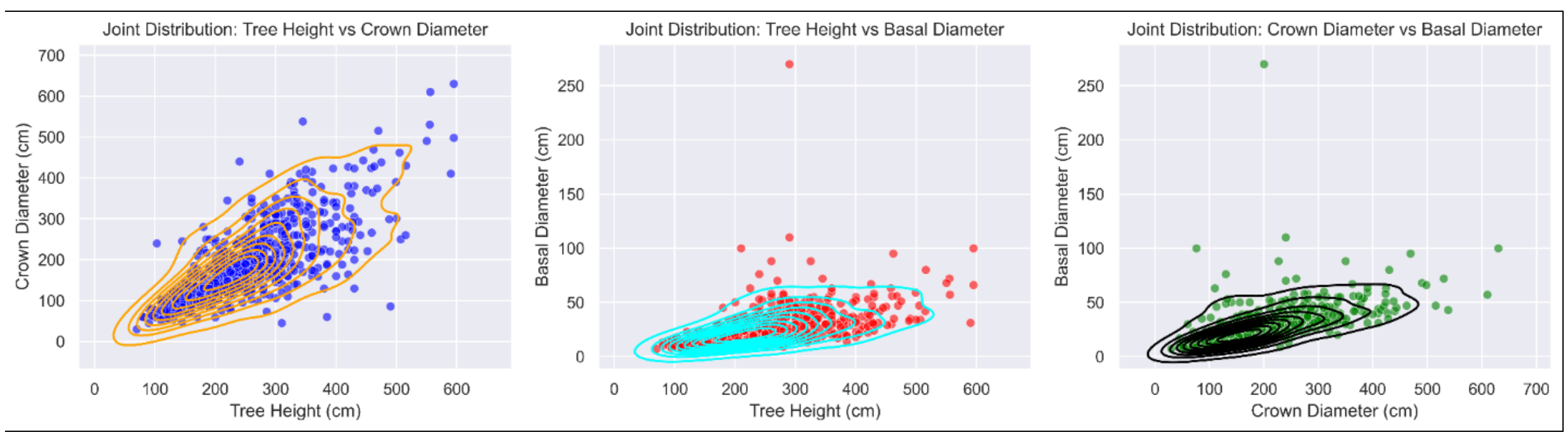

*Figure 10: Joint distributions of the various tree biophysical parameters in 2024*

## 5 Technical Validation

### 5.1 Tracking Individual Tree Growth

One use case of the Miti360 dataset is in tracking the growth of individual trees by monitoring the changes in tree crown sizes across time. To demonstrate this utility, we used the dataset's tree crown annotations to prompt Meta AI's Segment Anything Model (SAM), a promptable foundation model for semantic segmentation, to obtain tree crown masks. Figure 11 shows the tree crown masks overlayed on the tiles for images captured in March 2023 and August 2024 respectively.

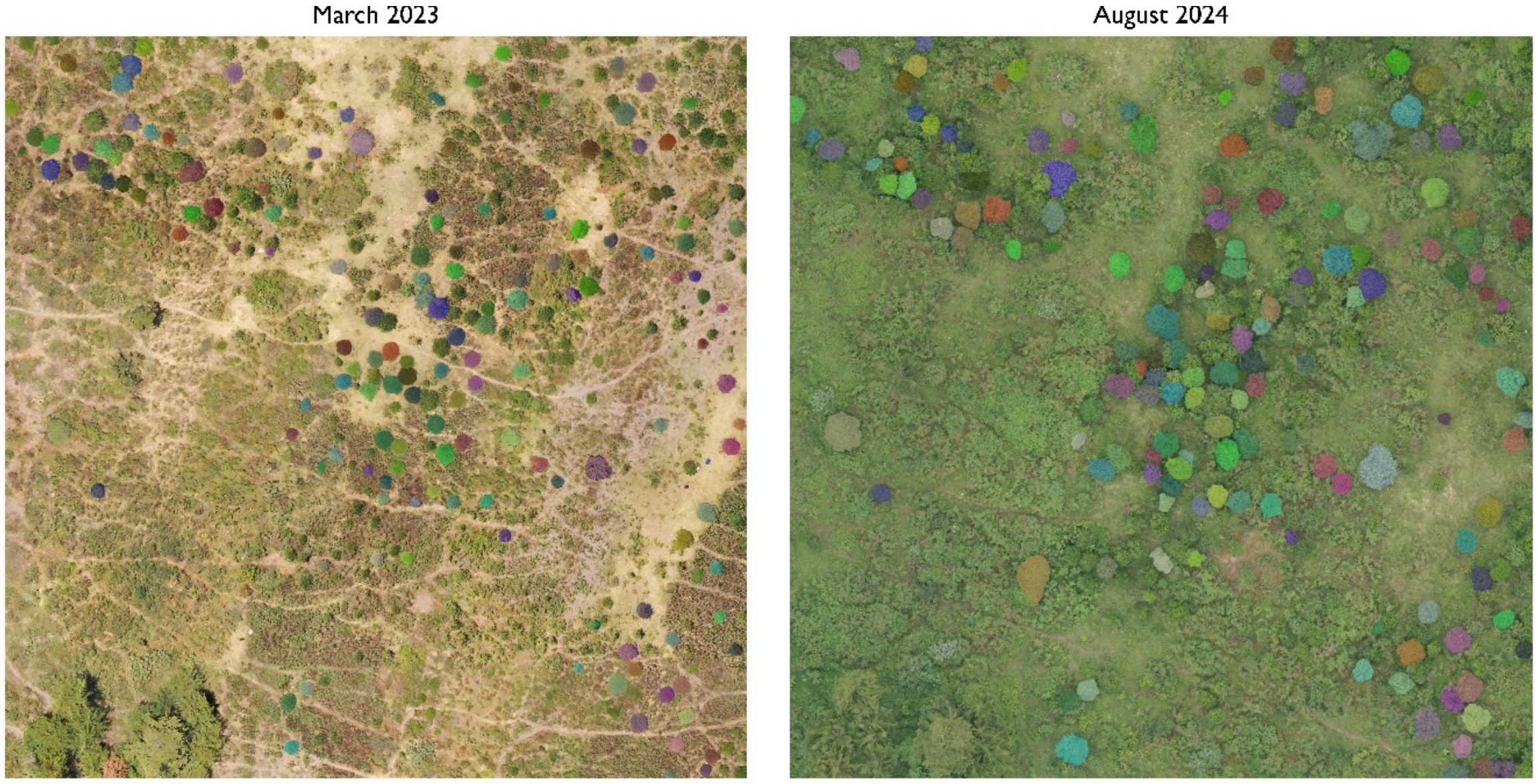

*Figure 11: Tree crown mask overlays showing change across time*

The sizes of the masks in real-world measurements were computed using the ground sampling distance, which corresponds to an area of ≈5.3 $cm^2$ per pixel. By finding the crown areas of 1585 trees, we quantified the change between March 2023 and August 2024. The box plot in Figure 12 shows a comparison between the crown areas at the two instances. As expected, there is an observable growth in the tree crown diameters with the mean values changing from ≈ 2 m to ≈ 3 m.

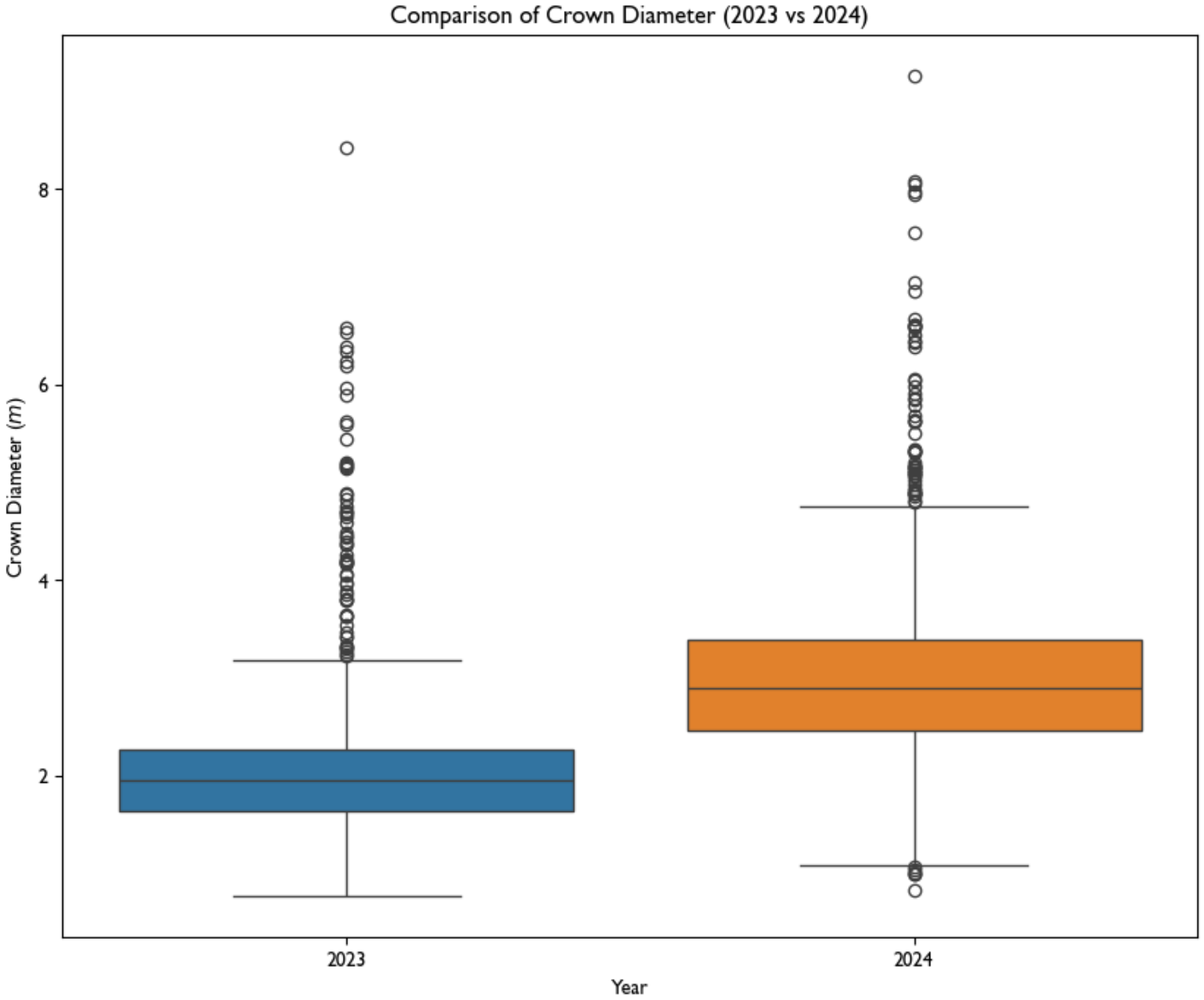

*Figure 12: Tree crown size comparison between 2024 and 2025*

## 5.2 Tree Crown Detection

The Miti360 dataset has nearly 62K tree crown annotations across each of the two orthophotos provided. These annotations can be used to train and finetune crown detection models which can be combined with semantic segmentation models to speed up the monitoring of tree growth, especially in young stands. Such an analysis would also be beneficial for developing tree growth models once data from more time points becomes available. We plan to keep growing the dataset to make this possible.

We finetuned the pretrained DeepForest tree detection model — a 32.1 M-parameter RetinaNet — on our dataset using a modest training setup of 10 epochs, a learning rate of 1e-4, a batch size of 100, and horizontal/vertical flip augmentations. Finetuning was conducted on a single GPU (CUDA) and saving the best checkpoint based on validation loss.

Quantitative results show dramatic improvements across three evaluation metrics (see Table 2) with the IoU threshold set at 0.4 as recommended by the publishers of the DeepForest package. Box precision improved from 2.8% to 22.7%, an ~8× gain, indicating the finetuned model fires far fewer spurious detections. Box recall surged from 20.5% to 89.6%, meaning the model now successfully locates nearly 9 in every 10 annotated trees, up from only 1 in 5. The mean IoU of detected trees also improved from 0.579 to 0.725, reflecting tighter bounding-box localisation around individual tree crowns.

*Table 2: Quantitative comparison between pretrained and finetuned deepforest*

| Metric | Pretrained | Finetuned |
|---|---|---|
| Box precision | 0.027 | 0.232 |
| Box Recall | 0.205 | 0.896 |
| Mean IoU | 0.579 | 0.725 |

Qualitatively, as shown in Figure 13, the pretrained model produces diffuse, poorly localised boxes that extend well into non-tree background regions. After the minimal finetuning, predictions latch tightly around individual canopies, demonstrating clear domain adaptation from the model's general-purpose training data to the aerial imagery characteristics of the Miti360 dataset.

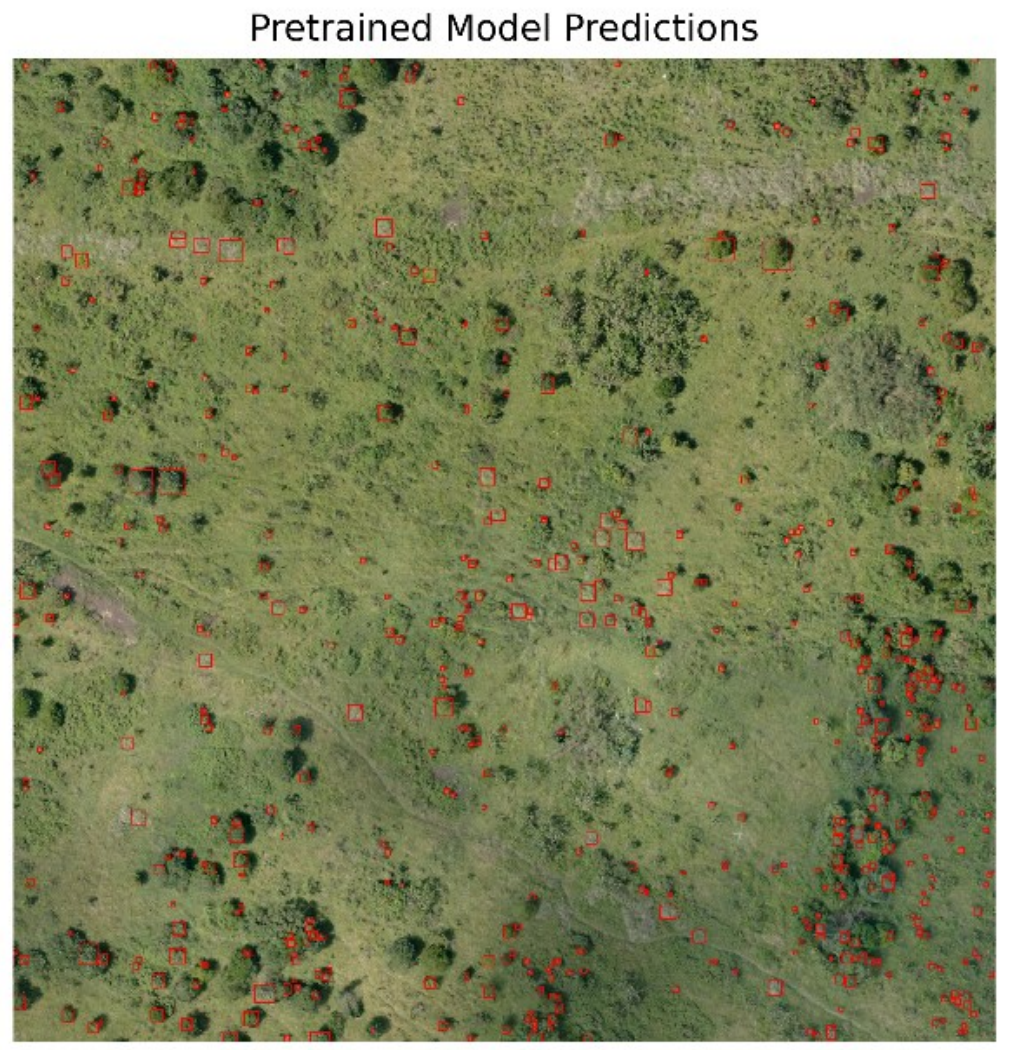

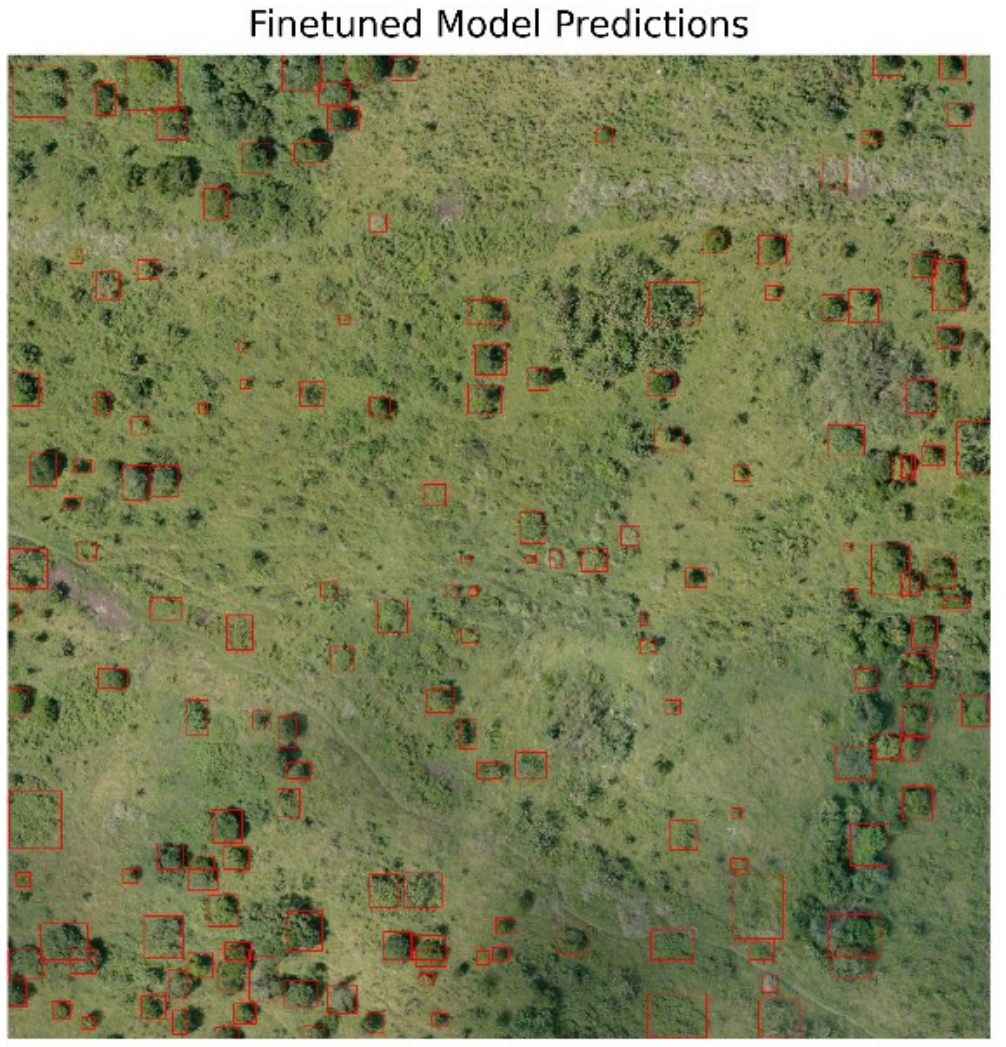

*Figure 13: Qualitative comparison between the pretrained and the finetuned crown detection model*

### 5.3 Effect of Precipitation on Tree Growth

Finally, we analyse the precipitation data from the TAHMO weather data to understand how rainfall influences tree growth. The data comprises precipitation records from 46 weather stations located within 100 km from the study area as seen in Figure 14.

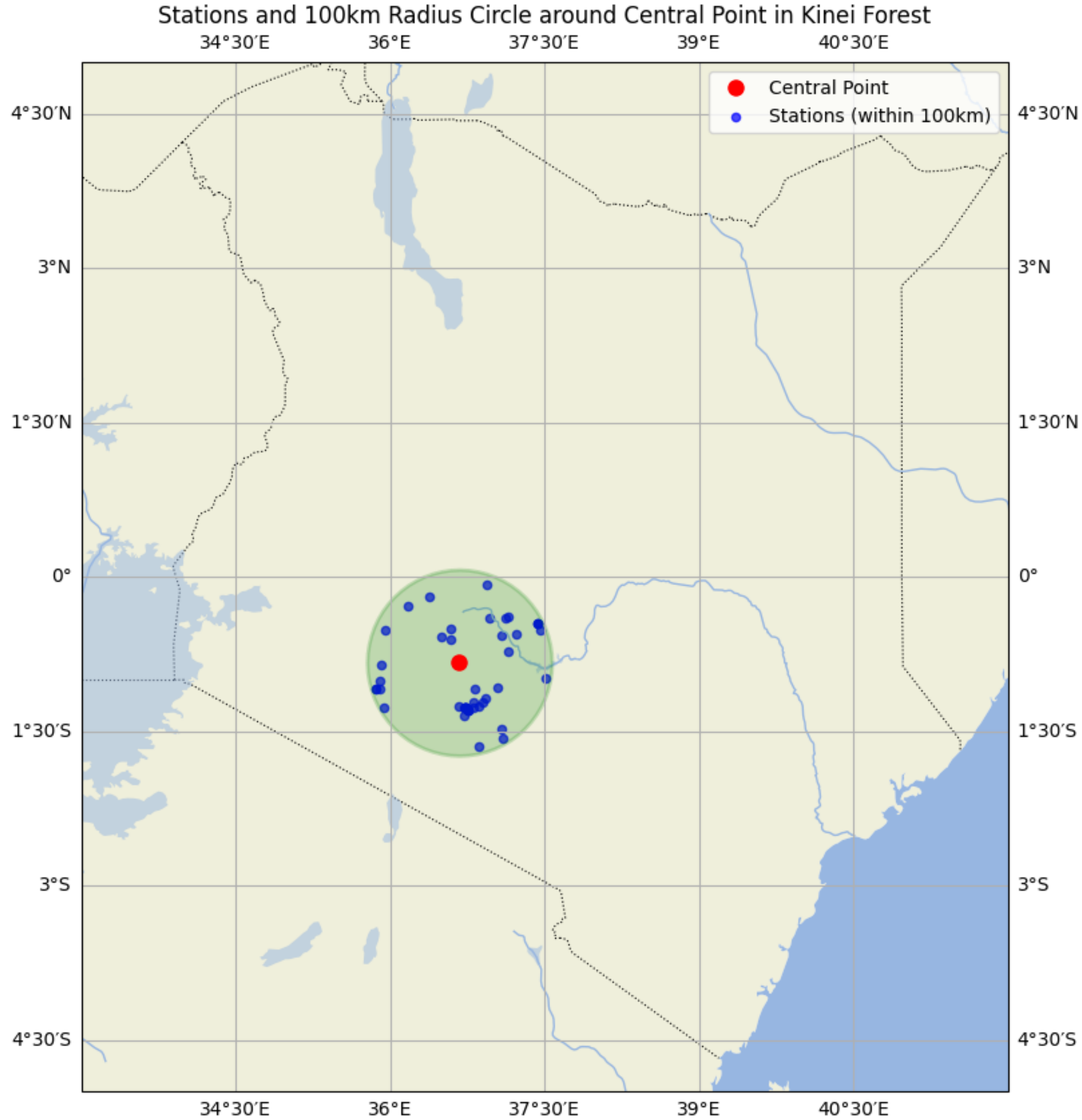

*Figure 14: A map showing the TAHMO weather stations located within 100 km from the study area*

Since none of these stations is within the forest and the nearest is approximately 28 km away, we performed interpolation to create a virtual rain gauge at the study area using inverse distance weighting (IDW) [49]. This method computes a weighted average of the surrounding stations (equations 1 and 2) where the assigned weight $w_i$ is proportional to its distance $d_i$ from the target raised to a power $p$ (we let $p$=2 in our case).

$$\hat{R} = \frac{\sum_{i=1}^{N} w_i R_i}{\sum_{i=1}^{N} w_i} \tag{1}$$

$$w_i = \frac{1}{d_i^p} \tag{2}$$

Once the rainfall values in the forest had been computed, the total rainfall and the duration of the longest dry spells in the periods intervening between the field data collection exercises as well as the change in the mean tree biophysical parameters over the same durations were calculated. The change in the mean parameters was computed as the difference between the mean values at the different sampling instances. These results are captured in Table 3 and Table 4.

*Table 3: Precipitation Metrics*

| Interval | Duration | Total Rainfall (mm) | Intensity (mm/month) | Max Dry Spell |
|---|---|---|---|---|
| Mar 23 – Aug 24 | 17 Months | 1156.08 | 72.26 | 15 |
| Aug 24 – Feb 25 | 6 Months | 407.56 | 57.40 | 10 |

*Table 4: Biophysical Parameter Progress Matrix*

| Parameter | Mean '23 | Mean '24 | Mean '25 | Δ Interval 1 | Rate 1 | Δ Interval 2 | Rate 2 |
|---|---|---|---|---|---|---|---|
| TH (cm) | 210.25 | 261.38 | 283.96 | 51.13 | 3.00 | 22.58 | 3.74 |
| CD (cm) | 150.81 | 198.37 | 210.37 | 47.56 | 2.79 | 12.00 | 1.99 |
| BD (cm) | – | 27.31 | 29.98 | – | – | 2.67 | 0.44 |

Since the two intervals over which the tree growth was observed are of highly unequal lengths (17 months vs 6 months), we used the normalised values (monthly rainfall and tree growth) for making comparisons. We notice that there was a significant decrease in monthly rainfall and reduction in the maximum dry spell between the two intervals leading to an increase in the crown diameter growth velocity. In contrast, tree height did not respond negatively to reduced rainfall intensity as expected but instead showed an increase in growth velocity from 3.00 mm/month to 3.74 mm/month. It was not mathematically possible to state whether changes in rainfall affected radial trunk growth because the basal diameter was not measured in 2023. Overall, our observation from the dataset is that changes in rainfall do not track linearly with tree growth rates.

## 6 Usage Notes

As various stakeholders engage in reforestation efforts, a key question is which trees should be planted, where they should be planted and when they should be planted. In preliminary work aimed at monitoring indigenous tree species at a 770-ha *reforested* stand in Kieni, Kenya [50], as well during our data collection exercises, we observed that some strategies employed include planting a wide variety of species with the hope that some would succeed. Miti360 will be useful in developing systems that power data driven decisions for stakeholders involved in the establishment and the maintenance of reforested stands. In many of Kenya's forests, plantations are often established with great assistance from farmers who are allowed to grow food crops within recently established stands. Our data will validate the contribution of these local farmers to reforestation efforts.

Miti360 can be used in varied ways to train and assess machine learning models. One example is to build models that can detect individual trees in the presence of dense shrubs that make it difficult for human beings to tell trees apart from shrubs. The labels of trees in the orthophotos will be useful for achieving that goal. One useful research angle we have pursued in the past is that of automating tree inventory using stereoscopic photogrammetry [44], [50]. With recent advances in deep learning and 3D computer vision, the stereoscopic images in Miti360 would be invaluable in developing better techniques for achieving the same goals. Regardless of the ways in which dataset may be used, we believe that all efforts directed towards developing novel techniques for forest monitoring tailored towards our African context will produce the greatest impact.

Overall, this data will serve stakeholders by enabling the development of machine learning systems capable of:

1. Increasing the efficiency of tree monitoring operations by quickly processing data collected by drones to determine individual tree counts.
2. Determining which species are growing well in a given area. This will be demonstrated by determining change in tree biophysical parameters over a period of one year.
3. Determining quantitative relationships between tree growth and weather.

## 7 Data Availability

The labelled dataset and all associated catalogues are available for download through our GitHub repository or the dataset's website.

## 8 Code Availability

The source code used to analyse the data is also accessible through our GitHub repository. Individual Jupyter notebooks showcasing various experiments with the data are stored in the "notebooks" folder in the repository.

## 9 Acknowledgements

This work was carried out with support from Lacuna Fund and Deutsche Gesellschaft für Internationale Zusammenarbeit (GIZ).

The views expressed herein do not necessarily represent those of Lacuna Fund, its Steering Committee, its funders, or Meridian Institute.

This work was supported in part by the Artificial Intelligence for Development (AI4D) program, with the financial support of the UK government's Foreign, Commonwealth, and Development Office (FCDO) and Canada's International Development Research Centre (IDRC).

This work was supported by a grant from the Swiss National Supercomputing Centre (CSCS) under project ID g164 on Alps.

We would like to acknowledge the domain expertise received from Integrated Forestry Consultancy and Management Services (IFCMS).

The authors gratefully acknowledge the Kenya Forestry Service for granting access to the study area and providing security to our team during the data collection.